\documentclass[10pt,twocolumn,letterpaper]{article}

\usepackage{cvpr}
\usepackage{times}
\usepackage{epsfig}
\usepackage{graphicx}
\usepackage{amsmath}
\usepackage{amssymb}
\usepackage{color}
\usepackage{float}
\usepackage{caption,subcaption}
\usepackage{hhline}

\usepackage[pagebackref=true,breaklinks=true,letterpaper=true,colorlinks,bookmarks=false]{hyperref}

\cvprfinalcopy 

\def\cvprPaperID{*} 

\ifcvprfinal\fi
\begin{document}

\title{Robust Face Detection via Learning Small Faces on Hard Images}

\author{Zhishuai Zhang$^{1}$~~~~~Wei Shen$^{1}$~~~~~Siyuan Qiao$^{1}$~~~~~Yan Wang$^{1}$~~~~~Bo Wang$^{2}$~~~~~Alan Yuille$^{1}$\\
Johns Hopkins University$^{1}$~~~Stanford University$^{2}$\\
{\tt\small \{zhshuai.zhang,shenwei1231,joe.siyuan.qiao,wyanny.9,wangbo.yunze,alan.l.yuille\}@gmail.com}
}

\maketitle

\begin{abstract}
Recent anchor-based deep face detectors have achieved promising performance, but they are still struggling to detect hard faces, such as small, blurred and partially occluded faces. A reason is that they treat all images and faces equally, without putting more effort on hard ones; however, many training images only contain easy faces, which are less helpful to achieve better performance on hard images. In this paper, we propose that the robustness of a face detector against hard faces can be improved by learning small faces on hard images. Our intuitions are (1) hard images are the images which contain at least one hard face, thus they facilitate training robust face detectors; (2) most hard faces are small faces and other types of hard faces can be easily converted to small faces by shrinking.
We build an anchor-based deep face detector, which only output a single feature map with small anchors, to specifically learn small faces and train it by a novel hard image mining strategy.
Extensive experiments have been conducted on WIDER FACE, FDDB, Pascal Faces, and AFW datasets
to show the effectiveness of our method.
Our method achieves APs of 95.7, 94.9 and 89.7 on \texttt{easy},
\texttt{medium} and \texttt{hard} WIDER FACE \texttt{val}
dataset respectively,
which surpass the previous state-of-the-arts, especially on the \texttt{hard} subset.
Code and model are available at \href{https://github.com/bairdzhang/smallhardface}{https://github.com/bairdzhang/smallhardface}.
\end{abstract}

\section{Introduction}
Face detection is a fundamental and important computer vision problem, which is critical for many face-related tasks, such as face alignment~\cite{cao2014face,xiong2013supervised}, tracking~\cite{kim2008face} and recognition~\cite{parkhi2015deep,schroff2015facenet}. Stem from the recent successful development of deep neural networks, massive CNN-based face detection approaches~\cite{hu2017finding,najibi2017ssh,Tang_2018_ECCV,zhang2017s,zhu2018seeing} have been
proposed and achieved the state-of-the-art performance. However, face detection remains a challenging task due to occlusion, illumination, makeup, as well as pose and scale variance, as shown in the benchmark dataset WIDER FACE~\cite{yang2016wider}.

Current state-of-the-art CNN-based face detectors attempt to address these challenges by employing more powerful backbone models~\cite{BaiFindingTF}, exploiting feature pyramid-style architectures to combine features from multiple detection feature maps~\cite{Tang_2018_ECCV}, designing denser anchors~\cite{zhu2018seeing} and utilizing larger contextual information~\cite{Tang_2018_ECCV}. These methods and techniques have been shown to be successful to build a robust face detector, and improve the performance
towards human-level for most images.

\begin{figure}[!t]
\includegraphics[height=2.85cm]{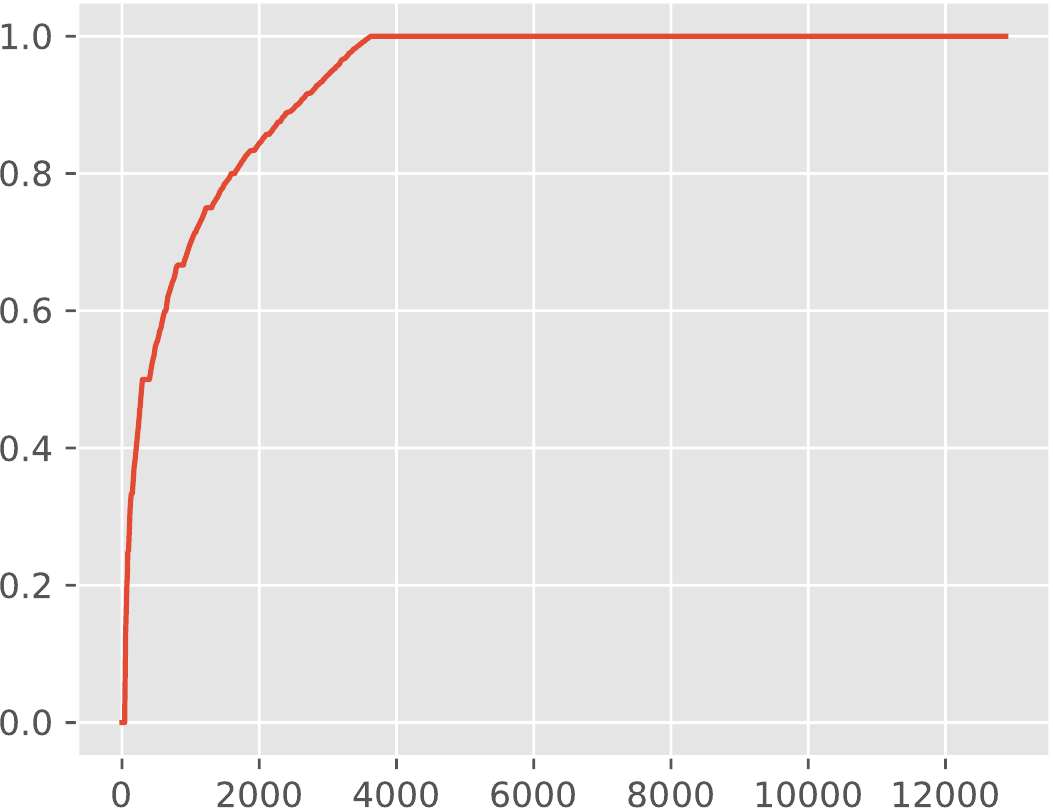}\hfill
\includegraphics[height=2.85cm]{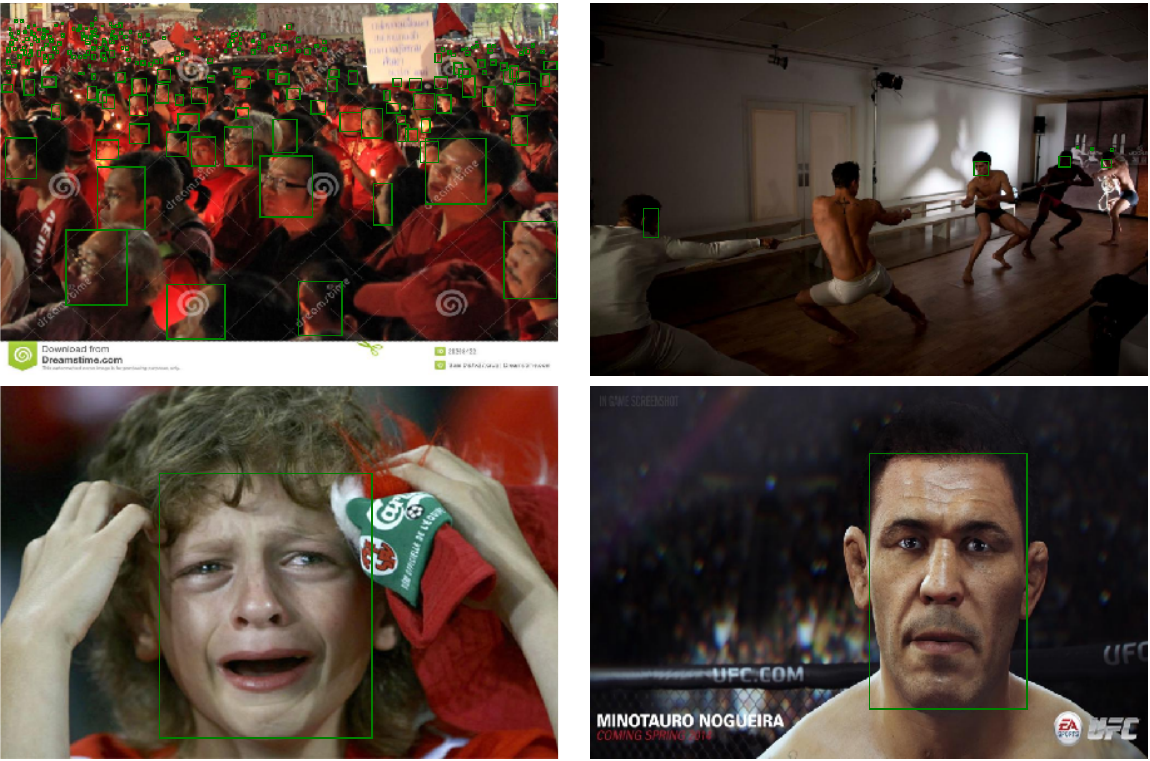}
\caption{Left: AP of each training image computed based on official SSH model, the x-axis
the is index of the training image, the y-axis is the AP for the corresponding image.
Upper right: hard training images. Lower right: easy training images.}
\label{fig:ssh_ap_train}
\end{figure}

In spite of their success for most images, an evident performance gap still exists especially for those hard images which contain small, blurred and partially occluded faces.
We realize that these hard images have become the main barriers for face detectors to achieve human-level detection performance.
In Figure~\ref{fig:ssh_ap_train}, we show that, even on the \texttt{train} set of WIDER
FACE, the official pre-trained SSH~\footnote{https://github.com/mahyarnajibi/SSH} still fails on some of the images with extremely
hard faces.
We show two such hard training images in the upper right corner in
Figure~\ref{fig:ssh_ap_train}.

On the other hand, most training images with easy faces can be almost perfectly detected (see the illustration in the right lower corner of Figure~\ref{fig:ssh_ap_train}).
As shown in left part of Figure~\ref{fig:ssh_ap_train}, over two thirds of the training images already obtained perfect detection accuracy, which indicates that those easy images are less useful towards training a robust face detector. To address this issue, in this paper, we propose a robust face detector by putting more training focus on those hard images.

This issue is most related to anchor-level hard example mining discussed in OHEM~\cite{shrivastava2016training}. However, due to the sparsity of ground-truth faces and positive
anchors, traditional anchor-level hard example mining mainly focuses on mining
hard negative anchors, and
mining hard anchors on well-detected images exhibits less effectiveness since there is no useful information that can be further exploited in these easy images.
To address this issue, we propose to mine hard examples at image level in parallel with anchor level. More specifically, we propose to dynamically assign
difficulty scores to training images during the learning process, which can determine whether an image is already well-detected or still useful for further training.
This allows us to fully utilize the images which were not perfectly detected to better facilitate the following learning process. We show this strategy can make our detector more robust towards hard faces, without involving more complex network architecture and computation overhead.

Apart from mining the hard images, we also propose to improve the detection quality by exclusively exploiting small faces.
Small faces are typically hard and have attracted extensive research attention~\cite{BaiFindingTF,hu2017finding,zhu2018seeing}. Existing methods aim at building a scale-invariant face detector to learn
and infer on both small and big faces, with multiple levels of detection features and anchors of different sizes.
Compared with these methods, our detector is more efficient since it is specially designed to aggressively leveraging the small faces during training.
More specifically, large faces are automatically ignored during training  due to our anchor design, so that the model can fully focus on the small hard faces.
Additionally, experiments demonstrate that this design effectively achieves improvements on detecting all faces in spite of its simple and shallow architecture.

\begin{figure*}[!ht]
\includegraphics[width=\linewidth]{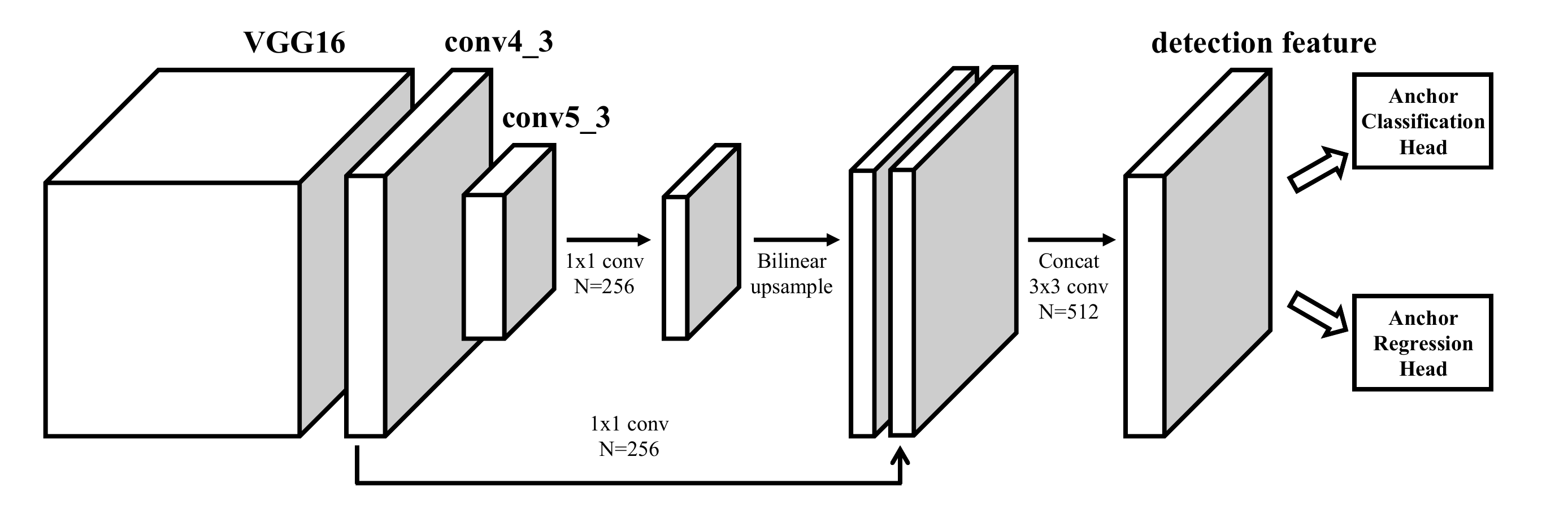}
\caption{The framework of our face detector. We take VGG16 as our backbone
CNN, and we fuse two layers (\texttt{conv4\_3} and \texttt{conv5\_3}) after dimension reduction
and bilinear upsampling, to generate the final detection feature map. Based on that,
we add a detection head for classification and bounding-box regression.}
\label{fig:arch}
\end{figure*}

To conclude, in this paper, we propose a novel face detector with the following
contributions:
\begin{itemize}
    \item We propose a hard image mining strategy, to improve the
    robustness of our detector to those extremely hard faces. This is done without
    any extra modules, parameters or computation overhead added on the existing detector.

    \item We design a single shot detector with only one detection feature map, which focuses on small faces with a
    specific range of sizes. This allows our model to be simple and focus on difficult
    small faces without struggling with scale variance.

    \item Our face detector establishes state-of-the-art performance on all popular face
    detection datasets, including WIDER FACE,
    FDDB, Pascal Faces, and AFW. We achieve 95.7, 94.9 and 89.7 on \texttt{easy}, \texttt{medium}
    and \texttt{hard} WIDER FACE \texttt{val} dataset. Our method also achieves APs of 99.00 and
    99.60 on Pascal Faces and AFW respectively, as well as a TPR of 98.7 on FDDB.
\end{itemize}

The remainder of this paper is organized as follows. In Section~\ref{sec:related_work}, we
discuss some studies have been done which are related to our paper. In Section~\ref{sec:method},
we dive into details of our proposed method, and we discuss experiment results and ablation
experiments in Section~\ref{sec:exp}. Finally, conclusions are drawn in Section~\ref{sec:conclusion}.

\section{Related work}\label{sec:related_work}
Face detection has received extensive research attention~\cite{li2014efficient,mathias2014face,viola2004robust}. With the
emergence of modern CNN~\cite{he2016deep,krizhevsky2012imagenet,simonyan2014very} and object detector~\cite{dai2016r,liu2016ssd,redmon2016you,ren2015faster,Zhang_2018_CVPR},
there are
many face detectors proposed to achieve promising performances~\cite{najibi2017ssh,tang2017multiple,Tang_2018_ECCV,wang2017face,wang2017detecting,zhang2017s}, by adapting general object detection framework into face detection domain. We briefly review hard example mining, face detection architecture, and anchor design \& matching.

\subsection{Hard example mining}
Hard example mining is an important strategy to improve model quality, and
has been studied extensively in image classification \cite{loshchilov2015online} and general object detection \cite{lin2018focal,shrivastava2016training}. The main idea is to find some hard positive and
hard negative examples at each step, and put more effort into training on those hard
examples~\cite{rowley1998neural,viola2001rapid}. Recently, with modern detection
frameworks proposed to boost the performance, OHEM~\cite{shrivastava2016training} and Focal loss~\cite{lin2018focal} have been proposed to select hard examples. OHEM computed the gradients of the networks by
selecting the proposals with highest losses in every minibatch; while Focal loss aimed at naturally putting more focus on hard and
misclassified examples by adding a factor to the standard cross entropy criterion.
However, these algorithms mainly focused on anchor-level or proposal-level mining. It cannot handle the
imbalance of easy and hard images in the dataset.
In our paper, we propose to exploit hard
example mining on image level, \ie hard image mining, to improve the quality of face detector on extremely hard faces. More specifically,
we assign difficulty scores to training images while training with an SGD mechanism,
and re-sample the training images to build a new training subset at the next epoch.

\subsection{Face Detection Architecture}
Recent state-of-the-art face detectors are generally built based on Faster-RCNN~\cite{ren2015faster}, R-FCN~\cite{dai2016r} or SSD~\cite{liu2016ssd}.
SSH~\cite{najibi2017ssh} exploited the RPN (Region Proposal Network) from Faster-RCNN to detect faces, by
building three detection feature maps and designing six anchors with different
sizes attached to the detection feature maps. S$^{3}$FD~\cite{zhang2017s} and
PyramidBox~\cite{Tang_2018_ECCV}, on the other hand,
adopted SSD as their detection architecture with six different detection feature maps.
Different from S$^3$FD, PyramidBox exploited a feature pyramid-style structure
to combine features from different detection feature maps. Our proposed method, on the other
hand, only builds single level detection feature map, based on VGG16,
for classification and bounding-box regression, which is both simple and effective.

\subsection{Anchor design and matching}
Usually, anchors are designed to have different sizes to detect objects with different
scales, in order to build a scale-invariant detector. SSD as well as its follow-up detectors S$^3$FD and PyramidBox,
had six sets of anchors with different
sizes, ranging from ($16\times16$) to ($512\times 512$), and their network architectures had six levels
of detection feature maps, with resolutions ranging from $\frac{1}{4}$ to $\frac{1}{128}$,
respectively. Similarly, SSH had the same anchor setting, and those anchors were
attached to three levels of detection
feature maps with resolutions ranging from $\frac{1}{8}$ to $\frac{1}{32}$. The difference
between SSH and S$^3$DF is that in SSH, anchors with two neighboring sizes shared the same
detection feature map, while in S$^3$DF, anchors with different sizes are attached to
different detection feature maps.

SNIP~\cite{Singh2017AnAO} discussed an alternative approach to handle scales. It
showed that CNNs are not robust to changes in scale, so training and testing on the same
scales of an image pyramid can be a more optimal strategy. In our paper, we exploit
this idea by limiting the anchor sizes to be ($16\times 16$), ($32\times 32$) and ($64\times 64$).
Then those faces with either too small or too big sizes will not be matched to any of the anchors, thus
will be ignored during the training and testing. By removing those large anchors with sizes larger than
($64\times 64$), our network focuses more on small faces which are potentially more difficult.
To deal with large faces, we use multiscale training
and testing to resize them to match our anchors.
Experiments show this design performs well on both small and big faces, although it
has fewer detection feature maps and anchor sizes.

\section{Proposed method}\label{sec:method}
In this section, we introduce our proposed method for effective face
detection. We first discuss the architecture of our detector in Section~\ref{sec:arch},
then we elaborate our hard image mining strategy in Section~\ref{sec:HIM}, as well as some other useful training techniques in Section~\ref{sec:train}.

\subsection{Single-level small face detection framework}\label{sec:arch}
The framework of our face detector is illustrated in Figure~\ref{fig:arch}. We use
VGG16 network as our backbone CNN, and combine \texttt{conv4\_3} and \texttt{conv5\_3} features,
to build the detection feature map with both low-level
and high-level semantic information. Similar to SSH~\cite{najibi2017ssh}, we apply
1$\times$1 convolution layers after \texttt{conv4\_3} and \texttt{conv5\_3} to reduce
dimension, and then apply a 3$\times$3 convolution layer on the
concatenation of these two dimension reduced features. The output feature of the
3$\times$3 convolution layer is the final detection feature map, which will be fed into
the detection head for classification and bounding-box regression.

The detection feature map has a resolution of $\frac{1}{8}$ of the original image (of size
$H\times W$).
We attach three anchors at each point in the grid as default face detection
boxes. Then we do classification and bounding-box regression on those $3\times\frac{H}{8}\times\frac{W}{8}$ anchors.
Unlike many other face detectors which build multiple feature maps
to detect face with a variant range of scales, inspired by
SNIP~\cite{Singh2017AnAO}, faces are trained and inferred with roughly the same scales.
We only have one detection feature map, with three sets of anchors attached to it.
The anchors have sizes of ($16\times16$), ($32\times32$) and ($64\times64$), and the
aspect ratio is set to be 1. By making this configuration, our network only trains and infers on small and medium size of faces; and we propose to handle large
faces by shrinking the images in the test phase. We argue that there is no speed
or accuracy degradation for large faces, since inferring on a tiny image (with short
side containing 100 or 300 pixels) is very fast, and the shrinked large face will still
have enough information to be recognized.

\begin{figure}[!t]
\includegraphics[width=\linewidth]{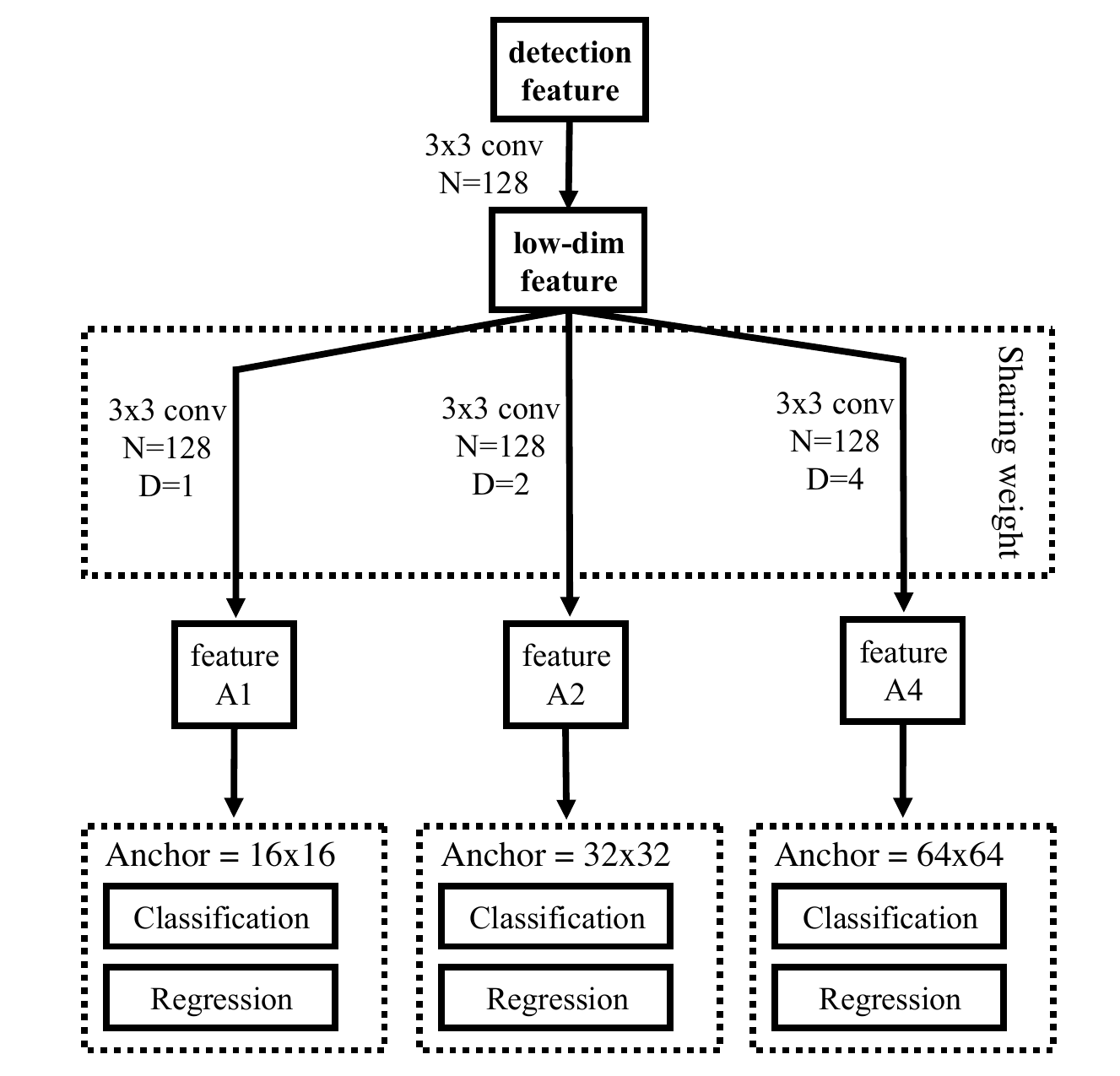}
\caption{The framework of our dilated detection head for classification and regression.
Based on the detection feature from the backbone CNN, we first
perform a dimension reduction to reduce the number of channels from 512 to 128.
Then we put three convolution layers with the shared weight, and different dilation rates,
to generate final detection and classification features.}
\label{fig:head}
\end{figure}
To handle the difference of anchor sizes attached to the same detection feature map, we propose a detection head which uses different dilation rates for anchors with
different sizes, as shown in Figure~\ref{fig:head}.
The intuition is that in order to detect faces with different sizes,
different effective receptive fields are required. This naturally
requires the backbone feature map to be invariant to scales. To this end,
we adopt different dilation rates for anchors with different sizes.
For anchors with size ($16\times 16$),
($32\times 32$) and ($64\times 64$), we use a convolution with kernel size of
$3$ and dilation rate of $1$, $2$ and $4$ to gather context features at
different scales. These three convolution layers share weights to
reduce the model size. With this design, the input of the $3\times 3$ convolution,
will be aligned to the same location of faces, regardless of the size of faces
and anchors. Ablation experiments show the effectiveness of this multi-dilation design.

\subsection{Hard image mining}\label{sec:HIM}
Different from OHEM discussed in Section~\ref{section:OHEM}, which selects proposals or anchors
with the highest losses,
we propose a novel
hard image mining strategy at image level. The intuition is that most images in the
dataset are very easy, and we can achieve a very high AP even on the \texttt{hard} subset
of the WIDER FACE \texttt{val}
dataset with our baseline model. We believe not all training images should be treated
equally, and well-recognized images will not help towards training a more robust
face detector. To put more attention on training hard images instead of easy ones,
we use a subset $\mathcal{D}'$ of all training images $\mathcal{D}$, to contain hard ones for
training. At the beginning of each epoch, we build $\mathcal{D}'$ based on
the difficulty scores obtained in the previous epoch.

We initially use all training images to train our model (\ie $\mathcal{D}'=\mathcal{D}$). This is due to the fact that our
initial ImageNet pre-trained model will only give random guess towards face detection. In this case, there is no easy image. In other words, every image is considered as hard image and fed to the network for training at the first epoch.
During the training procedure, we dynamically assign different difficulty scores to training images, which is defined
by the metric Worst Positive Anchor Score (WPAS):
\begin{equation*}
\text{WPAS}(I;\Theta)=\min_{a\in\mathcal{A}(I)^+}
\frac{\exp(l(I;\Theta)_{a,1})}{\exp(l(I;\Theta)_{a,1})+\exp(l(I;\Theta)_{a,0})}
\end{equation*}
where $\mathcal{A}(I)^+$ is the set of positive anchors for image $I$,
with IoU over 0.5 against ground-truth
boxes, $l$ is the classification logit and $l(I;\Theta)_{a,1}$,
$l(I;\Theta)_{a,0}$ are
the logits of anchor $a$ for image $I$ to be foreground face and background.
All images are initially marked as hard, and any image with WPAS greater than a threshold $th$
will be marked as easy image.

At the beginning of each epoch, we first randomly shuffle the training dataset to
generate the complete training list $\mathcal{D}=[I_{i_1},I_{i_2},\cdots,I_{i_n}]$ for the following
epoch of training. Then
given an image marked as easy, we remove it from $\mathcal{D}$ with
a probability of $p$.
The remaining training list $\mathcal{D}'=[I_{i_{j_i}},
I_{i_{j_2}},\cdots,I_{i_{j_k}}]$, which focuses more on hard images,
will be used for training at this epoch. Note that for
multi-GPU training, each GPU will maintain its training list $\mathcal{D}'$ independently.
In our experiments, we set the probability $p$ to be 0.7, and the threshold $th$ to be 0.85.

\subsection{Training strategy}\label{sec:train}
\subsubsection*{Multi-scale training and anchor matching}\label{sec:multiscale}
Since we only have anchors covering a limited range of face scales, we train our model by varying the sizes of training images. During the training phase, we resize
the training images so that the short side of the image contains $s$ pixels, where $s$ is
randomly selected from $\{400,800,1200\}$. We also set an upper bound of 2000 pixels to the long side of the image considering the GPU memory limitation.

For each anchor, we assign a label $\{+1, 0, -1\}$ based on how well it matches with any ground-truth face bounding box. If an anchor has an IoU (Intersection over Union) over
0.5 against a ground-truth face bounding
box, we assign $+1$ to that anchor. On the other hand, if the IoU against any ground-truth
face bounding box is lower than 0.3, we assign $0$ to that anchor. All other anchors will be given $-1$
as the label, and thus will be ignored in the classification loss.
By doing so, we only train on faces with designated scales. Those faces with
no anchor matching will be simply ignored, since we do not assign the anchor with largest
IoU to it (thus assign the corresponding anchor label $+1$) as Faster-RCNN does.
This anchor matching
strategy will ignore the large faces, and our model can put more
capacity on learning different face patterns on hard small faces
instead of memorizing the change in scales.

For the regression loss, all anchors with IoU greater than 0.3 against ground-truth faces will
be taken into account and contribute to the smooth $\ell1$ loss. We use
a smaller threshold (\ie 0.3) because (1) this will allow imperfectly matched
anchors to be able to localize the face, which may be useful during the
testing and (2) the regression task has less supervision since unlike
classification, there are no negative anchors for computing loss and the positive anchors are usually
sparse.
\subsubsection*{Anchor-level hard example mining}\label{section:OHEM}
OHEM has been proven to be useful for object detection and face detection in ~\cite{liu2016ssd,najibi2017ssh,shrivastava2016training}. During our
training, in parallel with our newly proposed hard image mining,
we also exploit the traditional hard anchor mining method to focus more on the
hard and misclassificed anchors. Given a training image with size $H\times W$,
there are $3\times\frac{H}{8}\times\frac{W}{8}$ anchors at the detection head,
and we only select 256 of them to be involved in computing the classification loss. For
all positive anchors with IoU greater than $0.5$ against ground-truth boxes, we
select the top 64 of them with lowest confidences to be recognized as face. After selecting positive
anchors, ($256-\#pos\_anchor$) negative anchors with highest face confidence are selected
to compute the classification loss as the hard negative anchors. Note that we only perform OHEM for classification loss, and we keep all anchors with IoU greater than 0.3 for computing regression loss, without selecting a subset based on either classification loss or bounding-box regression loss.
\subsubsection*{Data augmentation}\label{section:data_augmentation}
Data augmentation is extremely useful to make the model robust to light, scale changes and
small shifts~\cite{liu2016ssd,Tang_2018_ECCV}.
In our proposed method, we exploit cropping and photometric distortion as data
augmentation. Given a training image after resizing,
we crop a patch of it with a probability of
$0.5$. The patch has a height of $H'$ and a width of $W'$ which are independently drawn from
$\mathcal{U}(0.6H,H)$ and $\mathcal{U}(0.6W,W)$, where $\mathcal{U}$ is the uniform
distribution and $H$, $W$ are the height and width of the resized training image. All
ground-truth boxes whose centers are located inside the patch are kept.
After the random cropping, we apply photometric distortion following SSD by randomly modifying the brightness, contrast, saturation and
hue of the cropped image randomly.
\begin{figure*}[!t]
  \centering
  \begin{subfigure}[b]{0.33\linewidth}
    \centering\includegraphics[height=4.9cm]{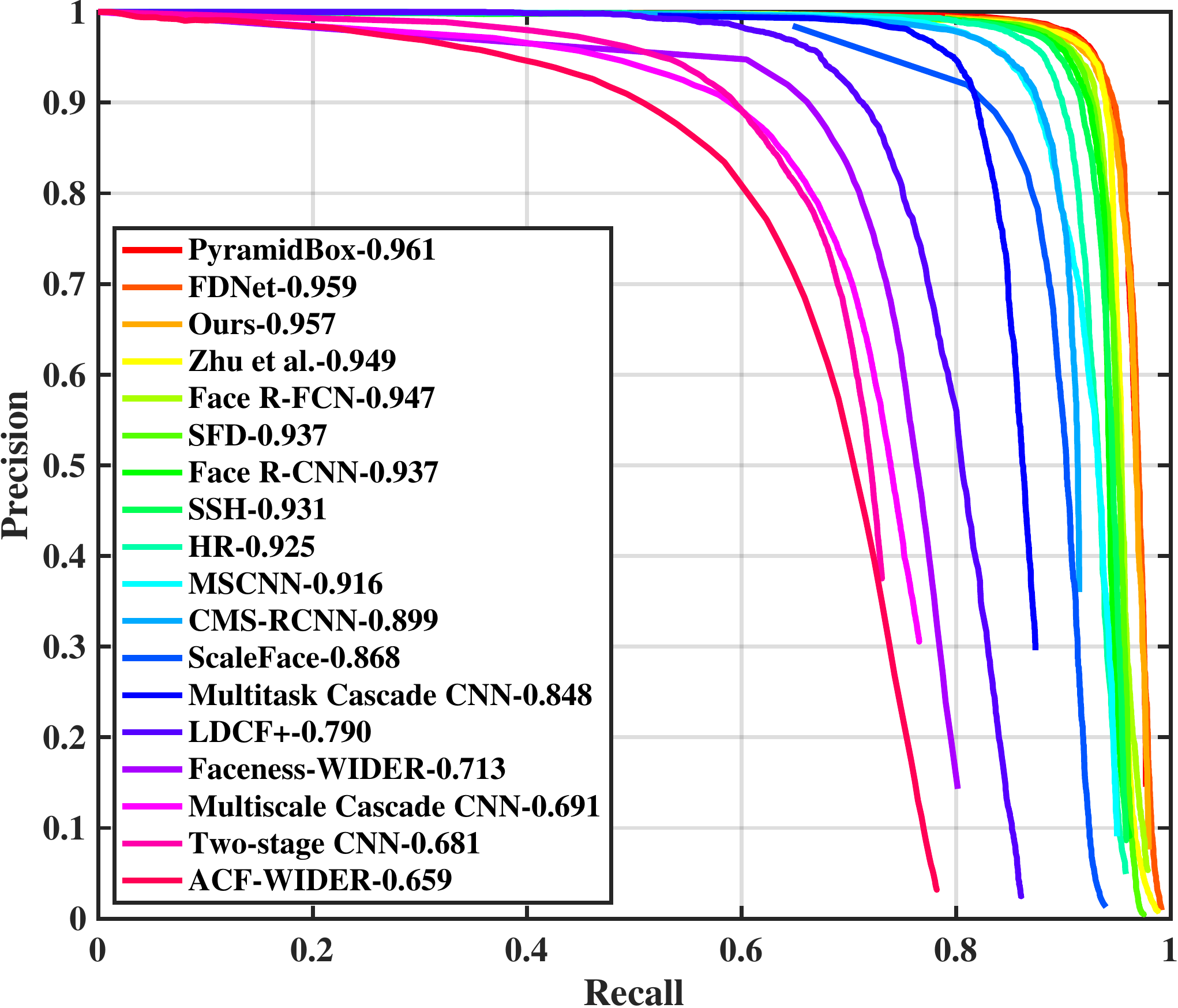}
    \caption*{\texttt{easy} subset}
  \end{subfigure}
   \begin{subfigure}[b]{0.33\linewidth}
    \centering\includegraphics[height=4.9cm]{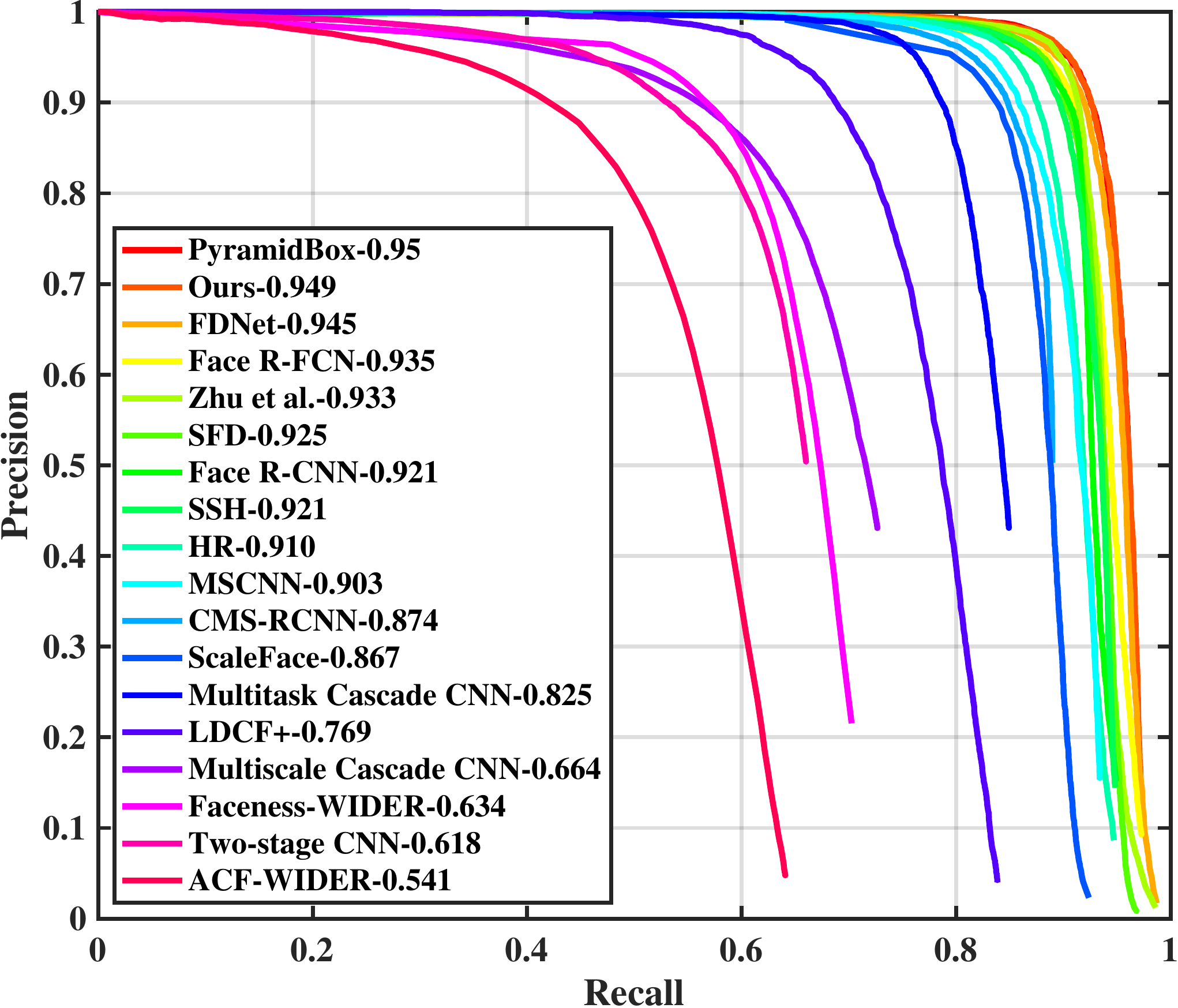}
    \caption*{\texttt{medium} subset}
  \end{subfigure}%
   \begin{subfigure}[b]{0.33\linewidth}
    \centering\includegraphics[height=4.9cm]{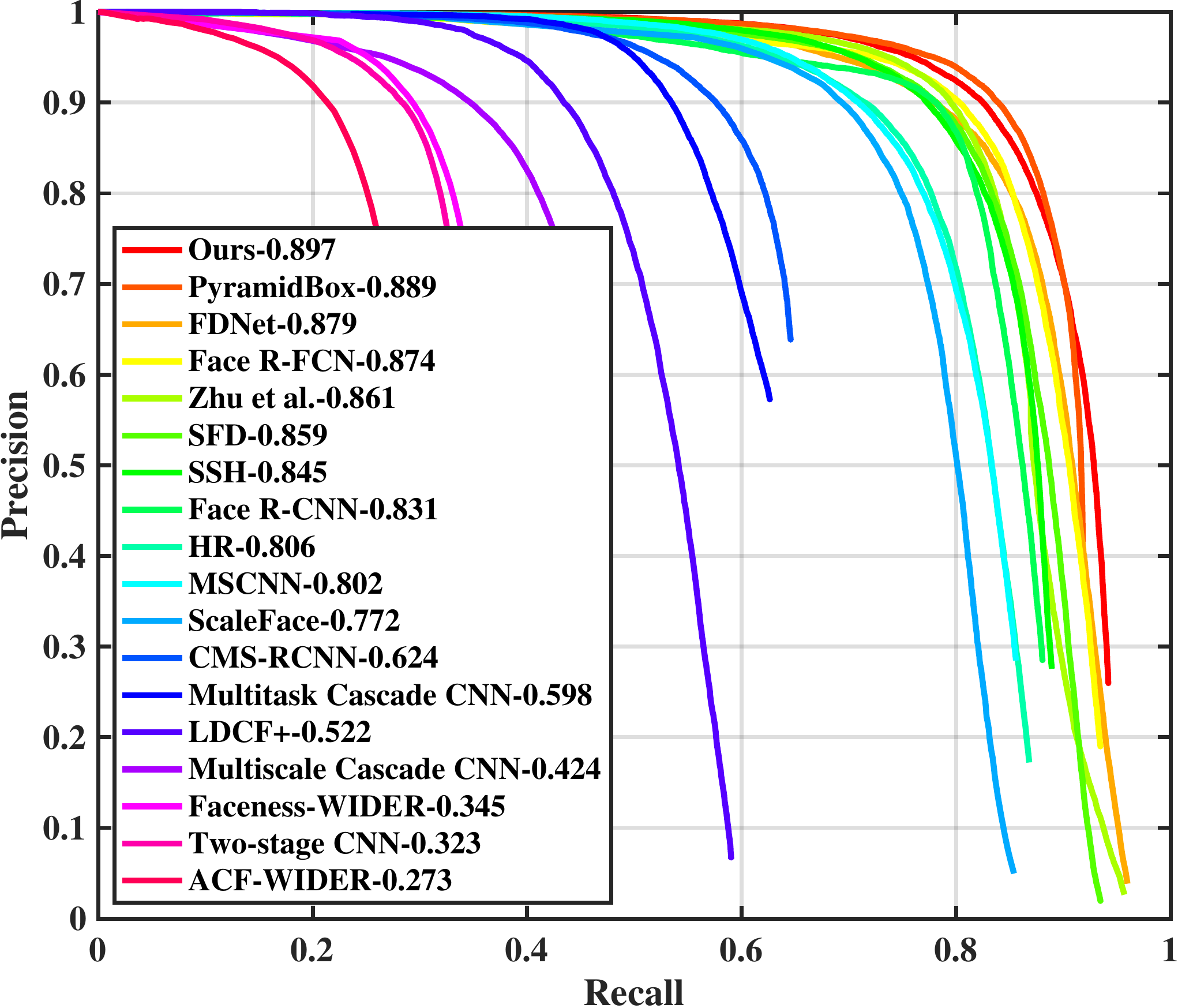}
    \caption*{\texttt{hard} subset}
  \end{subfigure}%
\caption{Precision-recall curve on WIDER FACE \texttt{val} dataset.}
\label{fig:perf_widerface}
\end{figure*}

\begin{figure}[!t]
    \centering\includegraphics[height=6.5cm]{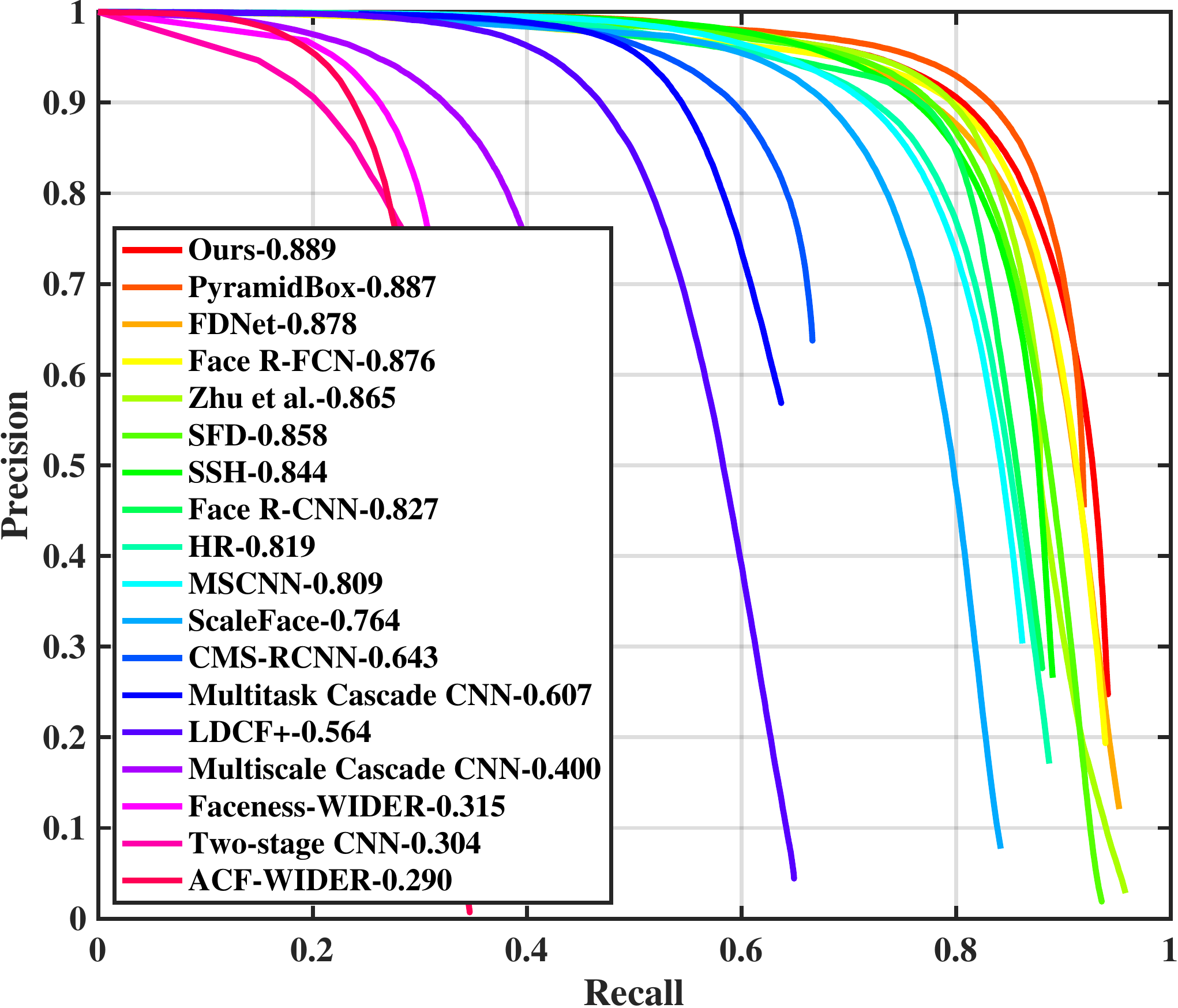}
\caption{Precision-recall curve on the \texttt{hard} subset of WIDER FACE \texttt{test} dataset.}
\label{fig:perf_widerface_test}
\end{figure}

\begin{figure*}[!t]
  \centering
\begin{subfigure}[b]{0.33\linewidth}
    \centering\includegraphics[height=4.85cm]{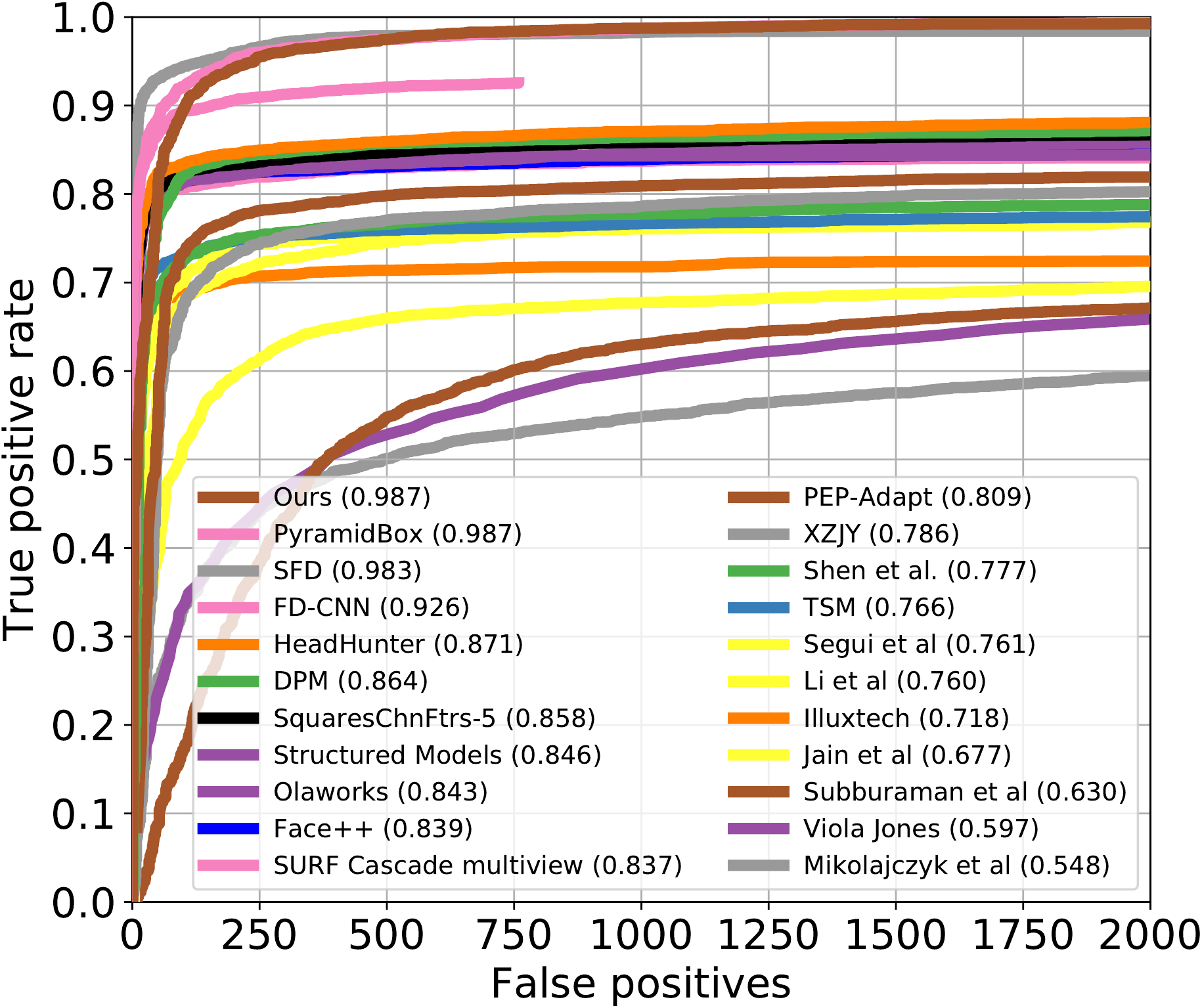}
    \caption{FDDB}\label{fig:fddb}
  \end{subfigure}
    \begin{subfigure}[b]{0.33\linewidth}
    \centering\includegraphics[height=4.85cm]{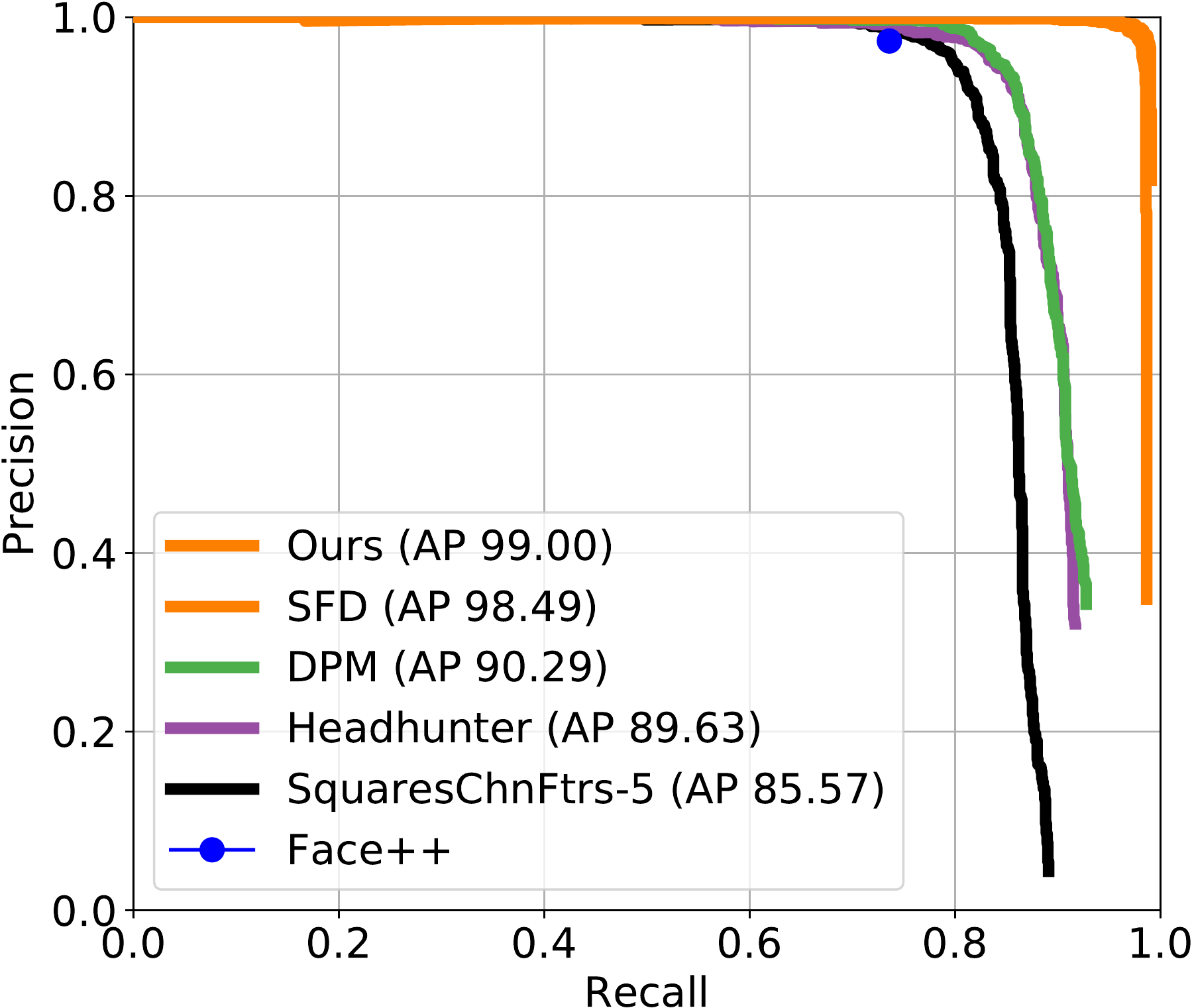}
    \caption{Pascal Faces}\label{fig:pascal}
  \end{subfigure}%
  \begin{subfigure}[b]{0.33\linewidth}
    \centering\includegraphics[height=4.85cm]{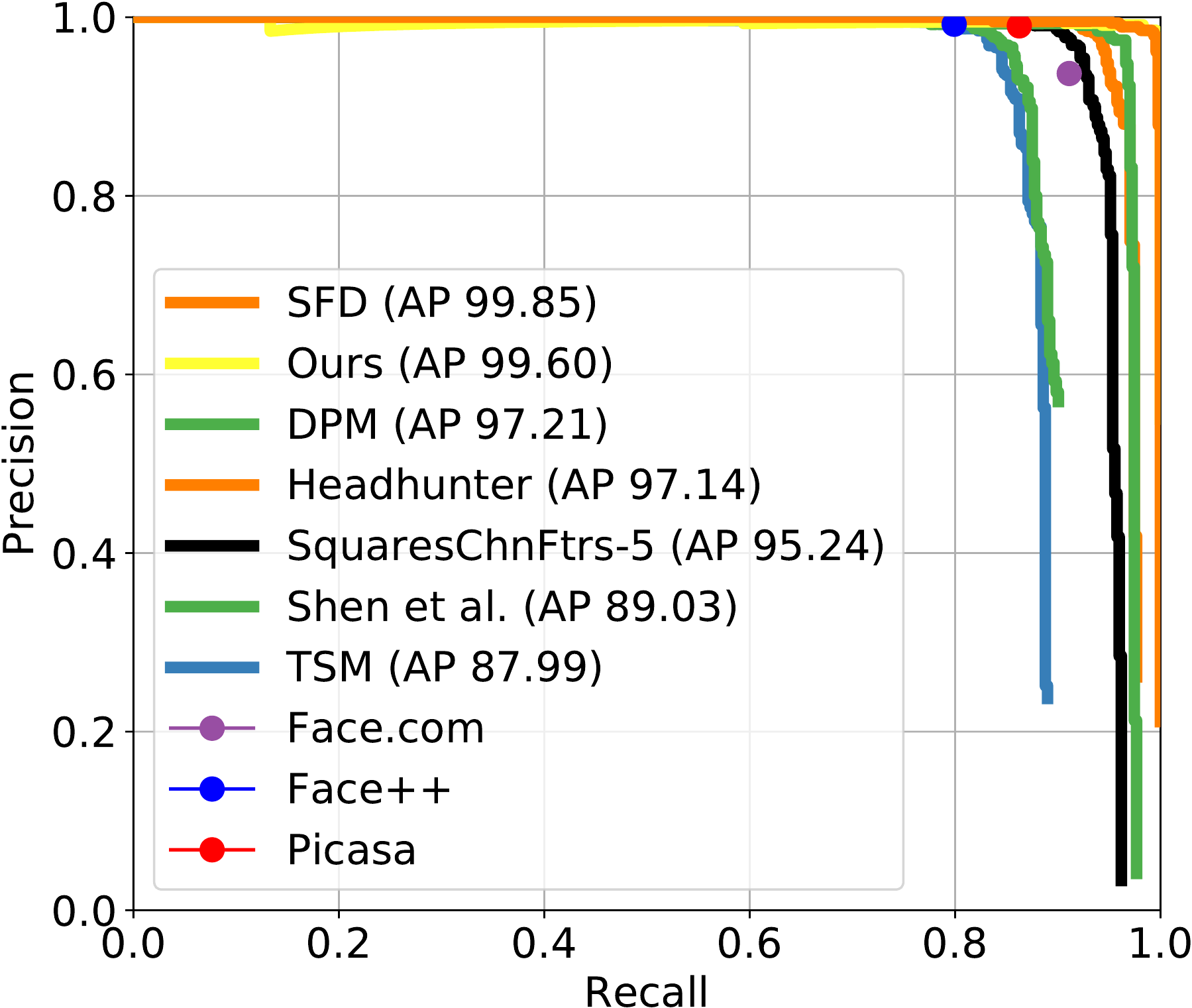}
    \caption{AFW}\label{fig:afw}
  \end{subfigure}
\caption{Performance compared with state-of-the-arts on other face datasets.}
\label{fig:perf_other}
\end{figure*}
\section{Experiments}\label{sec:exp}
To verify the effectiveness of our model and proposed method, we conduct extensive experiments
on popular face detection datasets, including WIDER FACE~\cite{yang2016wider}, FDDB~\cite{fddbTech}, Pascal Faces~\cite{yan2014face} and AFW~\cite{zhu2012face}.
It is worth noting that the training is only performed on the \texttt{train} set of WIDER FACE,
and we use the same model for evaluation on all these datasets without further fine-tuning.
\subsection{Experimental settings}
We train our model on the \texttt{train} set of WIDER FACE, which has 12880 images with
159k faces annotated. We flip all images horizontally, to
double the size of our training dataset to 25760.
For each training image, we first randomly resize it,
and then we use the cropping and photometric distortion data
augmentation methods discussed in Section~\ref{section:data_augmentation} to pre-process
the resized image. We use an ImageNet pre-trained
VGG16~\cite{krizhevsky2012imagenet} model to initialize our network backbone, and our newly
introduced layers are randomly initialized with Gaussian initialization. We train the
model with the itersize to be 2, for 46k iterations, with a learning rate of $0.004$, and
then for another 14k iterations
with a smaller learning rate of $0.0004$.
During training, we use 4 GPUs to simultaneously to compute the gradient and update the weight by
synchronized SGD with
Momentum~\cite{qian1999momentum}. The first two blocks of VGG16 are frozen during the training,
and the rest
layers of VGG16 are set to have a double learning rate.

Since our model is designed and trained on only small faces, we use a multiscale image
pyramid for testing to deal with faces larger than our anchors. Specifically, we resize
the testing image so that the short side contains 100, 300, 600, 1000 and 1400 pixels
for evaluation on WIDER FACE dataset. We also follow the testing strategies used in PyramidBox~\cite{Tang_2018_ECCV}\footnote{https://github.com/PaddlePaddle/models/blob/develop/fluid/PaddleCV/
face\_detection/widerface\_eval.py} such as horizontal flip and
bounding-box voting~\cite{gidaris2015object}.
\subsection{Experiment results}
\vspace{1ex}\noindent \textbf{WIDER FACE}
dataset includes 3226 images and 39708 faces labelled in the \texttt{val} dataset, with three subsets --
\texttt{easy}, \texttt{medium} and \texttt{hard}. In Figure~\ref{fig:perf_widerface}, we show the precision-recall (PR) curve and average precision (AP) for
our model compared with many other state-of-the-arts~\cite{BaiFindingTF,cai2016unified,najibi2017ssh,ohn2016boost,Tang_2018_ECCV,wang2017facercnn,wang2017face,wang2017detecting,wang2017rfcn,yang2014aggregate,yang2015facial,yang2016wider,yang2017face,zhang2018face,zhang2016joint,zhang2017s,zhu2018seeing,zhu2017cms} on these three subsets.
As we can see, our method achieves the best performance on the \texttt{hard} subset, and outperforms
the current state-of-the-art by a large margin.
Since the \texttt{hard} set is a super set of \texttt{small} and
\texttt{medium}, which contains all faces taller than 10 pixels, the performance
on \texttt{hard} set can represent the performance on the full testing
dataset more accurately.
Our performance on the \texttt{medium} subset is comparable to the
most recent state-of-the-art and the performance on the \texttt{easy} subset is a bit worse since our
method focuses on learning hard faces, and the architecture of our model
is simpler compared with other state-of-the-arts.

There is also a WIDER FACE \texttt{test} dataset with no annotations provided publicly. It contains 16097 images, and
is evaluated by WIDER FACE author team. We report the performance of our method at Figure~\ref{fig:perf_widerface_test} for the \texttt{hard} subset.

\vspace{1ex} \noindent \textbf{FDDB}
 dataset includes 5171 faces on a set of 2845 images, and we use our model trained on WIDER FACE \texttt{train} set to infer
on the FDDB dataset. We use the raw bounding-box result without fitting it into ellipse to
compute ROC.
We show the discontinuous ROC curve at Figure~\ref{fig:fddb} compared with~\cite{li2013learning,mathias2014face,shen2013detecting,Tang_2018_ECCV,viola2004robust,yan2014face,zhang2018face,zhang2017s,zhu2012face}, and our method achieves
the state-of-the-art performance of TPR=98.7\% given 1000 false positives.

\vspace{1ex} \noindent \textbf{Pascal Faces}
 dataset includes 1335 labeled faces on a set of 851 images extracted for the Pascal VOC dataset.
We show the PR curve at Figure~\ref{fig:pascal} compared with~\cite{mathias2014face,zhang2017s}, and our method achieves a new
the state-of-the-art performance of AP=99.0.

\vspace{1ex} \noindent \textbf{AFW}
dataset includes 473 faces labelled in a set of 205 images. As shown in Figure~\ref{fig:afw} compared with~\cite{mathias2014face,shen2013detecting,zhang2017s,zhu2012face}, our method achieves
state-of-the-art and almost perfect performance, with an AP of 99.60.
\subsection{Ablation study and diagnosis}
\subsubsection*{Ablation experiments}
In order to verify the performance of our single level face detector,
as well as the effectiveness of our proposed hard image mining, the
dilated-head classification and regression structure, we conduct various ablation experiments on the
WIDER FACE \texttt{val} dataset. All results are summarized in Table~\ref{tbl:ablation}.
\begin{table}[tb]
\centering
\begin{tabular}{|l|l|l|l|}
\hline
Method                       & \texttt{easy}&\texttt{medium} &\texttt{hard} \\ \hhline{|=|=|=|=|}
Baseline-Three          & 95.0 & 93.8  & 88.5 \\ \hhline{|=|=|=|=|}
Baseline-Single         & 95.1 & 94.2  & 89.1 \\ \hline
+ HIM                        & 95.4 & 94.8  & 89.6 \\ \hline
+ DH                         & 95.4 & 94.5  & 89.3 \\ \hline
+ DH + HIM                   & \bf{95.7} & \bf{94.9}  & \bf{89.7} \\ \hline
\end{tabular}
\caption{Ablation experiments. Baseline-Three is a face
detector similar to SSH with three detection feature maps.
Baseline-Single is our proposed detector with single detection feature map
shown in Figure~\ref{fig:arch}. HIM and DH represents hard image mining (Subsection~\ref{sec:HIM}) and dilated head
architecture (Figure~\ref{fig:head}).\label{tbl:ablation}}
\end{table}

From Table~\ref{tbl:ablation}, we can see that our single level baseline model can achieve performance comparable to
the current state-of-the-art face detector, especially on the hard subset.
Our model with single detection feature map performs better than the one with three detection
feature maps, despite its shallower
structure, fewer parameters and anchors.
This confirms the effectiveness
of our simple face detector with single detection feature map focusing on small faces.

We also separately verify our newly proposed hard image mining (HIM) and dilated head architecture (DH) described in Subsection~\ref{sec:HIM} and Figure~\ref{fig:head}
respectively.
HIM can improve the performance on hard subset significantly without involving more complex network
architecture nor computation overhead. DH itself can also boost the performance, which
shows the effectiveness of designing larger convolution for larger anchors.
Combining HIM and DH together can improve further towards the state-of-the-art performance.
\subsubsection*{Diagnosis of hard image mining}
\begin{figure}[!tb]
\includegraphics[width=\linewidth]{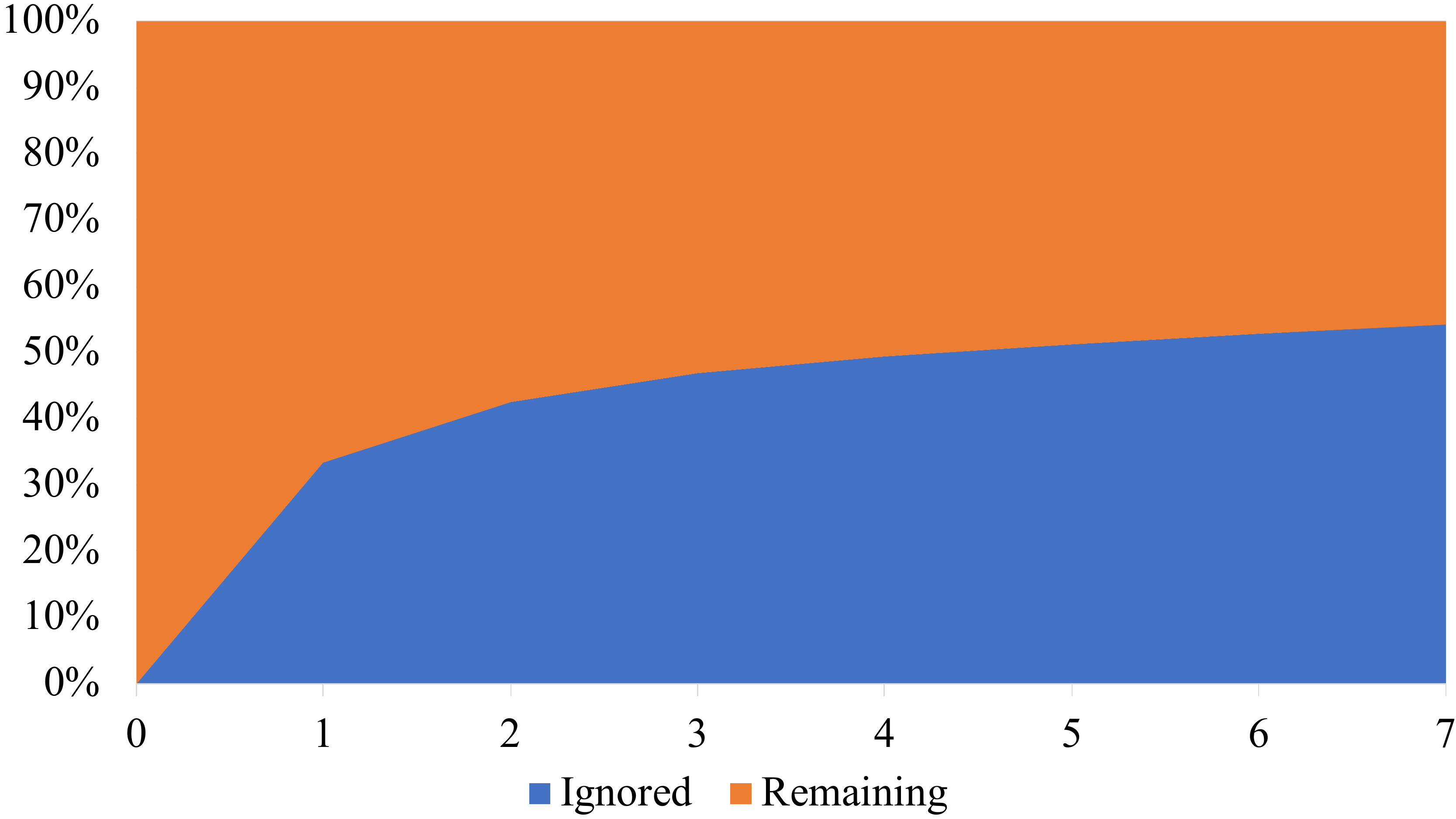}
\caption{Ratio of ignored images. X-axis is the epoch index and y-axis is the proportion
of ignored training images by hard image mining.}
\label{fig:ratio}
\end{figure}

We investigate the effects of our hard image mining mechanism. We show the ratio of $|\mathcal{D}'|$ and $|\mathcal{D}-\mathcal{D}'|$ (\ie the ratio of the number of selected training images to
the number of ignored training images) in Figure~\ref{fig:ratio} for each
epoch. We can see that at the
first epoch, all training images are used to train the model. Meanwhile, as the training process continues, more and more training images will be ignored. At the last epoch, over a half images will be ignored and thus will not be included in $\mathcal{D}'$.

\subsubsection*{Diagnosis of data augmentation}
We investigate the effectiveness of the photometric distortion
as well as the cropping mechanisms as discussed in
Subsection~\ref{section:data_augmentation}. The ablation results evaluated on WIDER
FACE \texttt{val} dataset are shown in
Table~\ref{tbl:da}. Both photometric distortion and cropping can contribute to a more
robust face detector.
\begin{table}[ht]
\centering
\begin{tabular}{|l|l|l|l|l|}
\hline
PD & Crop & \texttt{easy} & \texttt{medium} & \texttt{hard} \\ \hline
No  & No    & 94.5 & 93.9   & 88.4 \\ \hline
Yes  & No    & 94.5 & 94.1   & 88.8 \\ \hline
Yes  & Yes    & 95.1 & 94.2   & 89.1 \\ \hline
\end{tabular}
\caption{Diagnosis of data augmentation. PD indicates photometric distortion.
All entries are based on our Baseline-SingleLevel configuration without
HIM and DH.}\label{tbl:da}
\end{table}

\subsubsection*{Diagnosis of multi-scale testing}
Our face detector with one detection feature map is design for small face detection, and our anchors
are only capable of capturing faces with sizes ranging from ($16\times 16$) to
($64\times 64$). As a result, it is critical to adopt multi-scale testing to deal with large faces.
Different from SSH, S$^3$FD and PyramidBox, our testing pyramid includes some extreme
small scales (\ie short side contains only 100 or 300 pixels). In Table~\ref{tbl:scale},
we show the effectiveness of these extreme small scales to deal with easy and large images.
Our full evaluation resizes the image so that the short side contains 100, 300, 600, 1000
and 1400 pixels respectively, to build an image pyramid. We diagnose the impact of the
extra small scales (\ie 100 and 300) by removing them from the image pyramid.

\begin{table}[ht]
\centering
\begin{tabular}{|l|l|l|l|}
\hline
Testing Scales                  & \texttt{easy}& \texttt{medium} & \texttt{hard} \\ \hhline{|=|=|=|=|}
{[}600, 1000, 1400{]}           & 78.2     & 85.7      & 86.1     \\ \hline
{[}300, 600, 1000, 1400{]}      & 91.3     & 92.6      & 88.8     \\ \hline
{[}100, 300, 600, 1000, 1400{]} & 95.7 & 94.9  & 89.7 \\ \hline
\end{tabular}
\caption{Diagnosis of multi-scale testing.\label{tbl:scale}}
\end{table}

As shown in Table~\ref{tbl:scale}, the extra small scales are crucial to detect
easy faces. Without resizing the short side to contain 100 and 300 pixels, the performance
on \texttt{easy} subset is only $78.2$, which is even lower than the performance on \texttt{medium}
and \texttt{hard} which contain much harder faces. We will show in the next subsection that
these extra small scales ($100$ and $300$) lead to negligible computation overhead, due to the
lower resolution.

\subsubsection*{Diagnosis of accuracy/speed trade-off}
\begin{table}[!t]
\centering
\begin{tabular}{|l|l|l|l|l|l|}
\hline
Method     & MS  & HF  & Time & G-Mem & AP-h   \\ \hhline{|=|=|=|=|=|=|}
SSH        & Yes & No  & 1.00 & 6.1        & 84.5 \\ \hline
S$^3$FD       & Yes & Yes & 1.34 & 6.2        & 85.2 \\ \hline
PyramidBox & Yes & Yes & 2.24 & 11.9       & 88.9 \\ \hline
Ours$^*$       & Yes$^*$ & Yes  & 1.59 & \bf{5.3}        & 86.1    \\ \hline
Ours       & Yes & No  & \bf{0.84} & \bf{5.3}        &  89.3\\ \hline
Ours       & Yes & Yes & 1.70 & \bf{5.3}        &  \bf{89.7}    \\ \hline
\end{tabular}
\caption{Diagnosis of inference speed. MS and HF indicate multi-scale testing and
horizontal flip; Time is the inference time (in second) for a single image;
G-Mem is the GPU memory usage in gigabyte;
AP-h is the average precision on the \texttt{hard} subset of WIDER FACE
\texttt{val} set. Ours$^*$ indicates our detector without extra small scales.
All entries are evaluated with a single nVIDIA Titan X (Maxwell).\label{tbl:speed}}
\end{table}
We evaluate the speed of our method as well as some other popular face detectors in
Table~\ref{tbl:speed}. For fair comparison, we run all methods on the same machine,
with one Titan X (Maxwell) GPU, and Intel Core i7-4770K 3.50GHz. All methods
except for PyramidBox are based on Caffe1 implementation, which is compiled with CUDA 9.0 and CUDNN 7.
For PyramidBox, we follow the official fluid code and the default configurations\footnote{https://github.com/PaddlePaddle/models/blob/develop/fluid/PaddleCV/\\
face\_detection/widerface\_eval.py}. We
use the officially built PaddlePaddle with CUDA 9.0 and CUDNN 7\footnote{\texttt{pip install paddlepaddle-gpu}}.

For SSH, S$^3$FD and Pyramid, we use the official inference code and configurations. For SSH,
we use multi-scale testing with the short side containing 500, 800, 1200 and 1600
pixels, and for S$^3$FD, we execute the official evaluation code with both multi-scale
testing and horizontal flip. PyramidBox takes a similar testing configuration as S$^3$FD.
As shown in Table~\ref{tbl:speed}, our detector can
outperform SSH, S$^3$FD and PyramidBox significantly with a smaller inference time. Based on that,
using horizontal flip can further improve the performance slightly.
In terms of GPU memory usage, our method uses only a half of what PyramidBox occupies,
while achieving better performance.

Ours$^*$ in Table~\ref{tbl:speed} indicates our method without extra small scales in
inference, \ie, evaluated with scales {[}600, 1000, 1400{]}. It is only $6.5\%$ faster
than evaluation with {[}100, 300, 600, 1000, 1400{]} (1.59 compared with 1.70). This
proves that although our face detector is only trained on small faces, it can perform
well on large faces, by simply shrinking the testing image with negligible computation
overhead.

\section{Conclusion}\label{sec:conclusion}
To conclude, we propose a novel face detector to focus on learning small faces on hard images, which
achieves the state-of-the-art performance on all popular face detection datasets. We propose a hard
image mining strategy by dynamically assigning difficulty scores to training images, and re-sampling
subsets with hard images for training before each epoch.
We also design a single shot face detector with only one detection feature map, to train and test
on small faces. With these designs, our model can put more
attention on learning small hard faces instead of memorizing change of scales.
Extensive experiments and ablations have
been done to show the effectiveness of our method, and our face detector achieves the
state-of-the-art performance on all popular face detection datasets, including WIDER FACE, FDDB, Pascal Faces and AFW. Our face detector also enjoys faster multi-scale inference
speed and less GPU memory usage.
Our proposed method are flexible and can be applied to other backbones and tasks,
which we remain as future work.

{\small
\bibliographystyle{ieee}
\bibliography{egbib}
}

\end{document}